\documentclass[12pt]{article}

\usepackage{sbc-template}
\usepackage{abntex2cite}
\usepackage{graphicx,url}
\usepackage[utf8]{inputenc}  
\usepackage[T1]{fontenc} 
\usepackage{textcomp}
\usepackage{ae} 
\usepackage{amsmath}
\usepackage{amsfonts}
\usepackage{amssymb}
\usepackage{array}
\usepackage{blindtext}
\usepackage{dcolumn}
\usepackage{verbatim}
\usepackage{enumitem}
\usepackage{parskip}
\usepackage[table,xcdraw]{xcolor}
\usepackage{lipsum}
\usepackage{pifont}
\usepackage{afterpage}
\usepackage{imakeidx}
\usepackage{setspace}
\usepackage{pbox}
\usepackage{wrapfig}
\usepackage{titlesec}
\usepackage{dsfont} 
\usepackage{ulem}
\usepackage{bbm}
\usepackage{MnSymbol}
\usepackage{color}
\usepackage{makecell}
\usepackage{footnote}
\usepackage{tablefootnote}

\definecolor{codegreen}{rgb}{0,0.6,0}
\definecolor{codegray}{rgb}{0.5,0.5,0.5}
\definecolor{codepurple}{rgb}{0.58,0,0.82}
\definecolor{backcolour}{rgb}{0.95,0.95,0.92}

\title{A Summary Description of the A2RD Project}

\author{Juliao Braga\inst{1,2}, Joao Nuno Silva\inst{1}, Patricia Takako Endo\inst{3,4}, Nizam Omar\inst{2}}

\address{IST - INESC ID, University of Lisboa, Portugal
\nextinstitute
Universidade Presbiteriana Mackenzie (UPM)
\nextinstitute
Universidade de Pernambuco (UPE), Brazil
\nextinstitute
Dublin City University (DCU), Ireland
\email{\{juliao.braga,joao.n.silva\}@tecnico.ulisboa.pt}
\email{patricia.endo@upe.br,nizam.omar@mackenzie.br}
}

\begin{document} 

\maketitle

\begin{abstract}
This paper describes the Autonomous Architecture Over Restricted Domains  project. It begins with the description of the context upon which the project is focused, and in the sequence describes the project and implementation models. It finish by presenting the environment conceptual model, showing where stand the components, inputs and facilities required to interact among the intelligent agents of the various implementations in their respective and restricted, routing domains (Autonomous Systems) which together make the Internet work.
\end{abstract}

\section{Introduction}

\begin{figure}[!ht]
\centering
\includegraphics[scale=0.6]{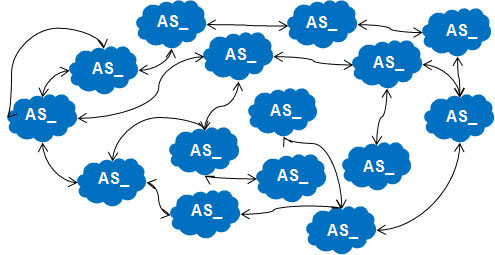}
\caption{How ASs build the Internet: $\_$ is some $i$, where $ 0 < i < 4.294.967.296 \simeq (2^{32})$}
\label{fig:000internet}
\end{figure}

\textit{Autonomous System} (AS) is the name given to the networks that making up the Internet \cite{RFC1930:1996}. This cluster of ASes interconnected also called routing domains or more commonly the Internet, can be represented as in Figure \ref{fig:000internet}.

The ASes establish interconnections through a protocol called the \textit{Border Gateway Protocol} (BGP) \cite{RFC4271}. BGP is a complex protocol that requires a lot of knowledge from the administrators of an AS. In addition to the complexity of BGP, one must add the complexity of Internet governance, which is partially visible in Figure \ref{fig:InternetEcosystem}.

\afterpage{
\begin{figure}[!ht]
\centering
\includegraphics[scale=0.37]{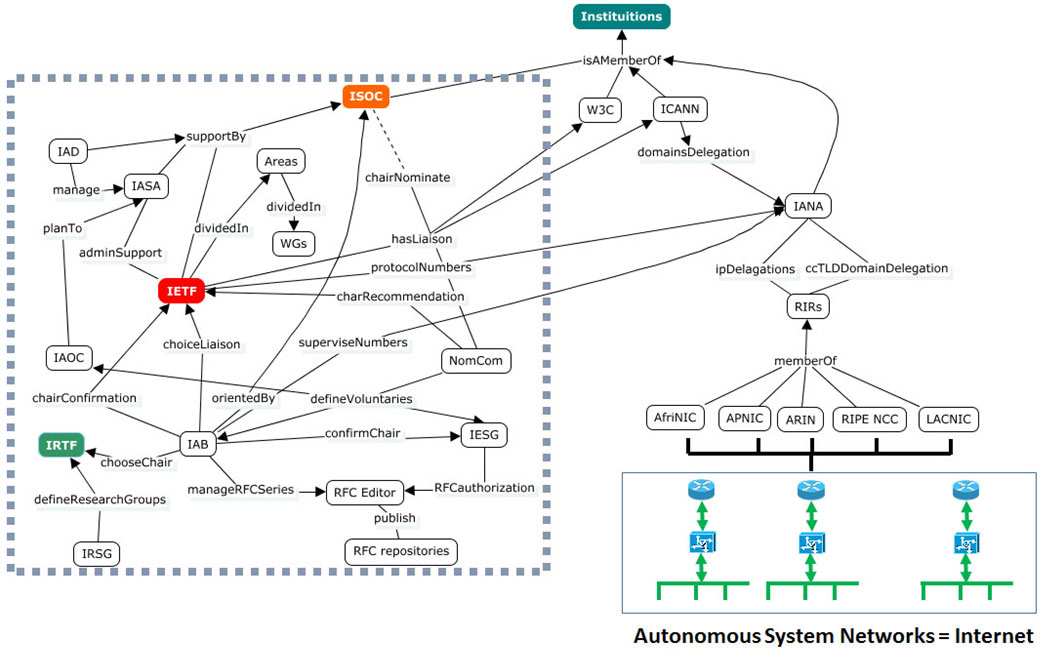}
\caption[Caption for LOF]{The Internet Infrastructure Ecosystem. On 02/03/2018 there were 59.959 ASes present in the Internet Routing Table\footnotemark}
\label{fig:InternetEcosystem}
\end{figure}
\footnotetext{http://thyme.rand.apnic.net/current/data-summary}
}

\begin{figure}[!ht]
\centering
\includegraphics[scale=0.3]{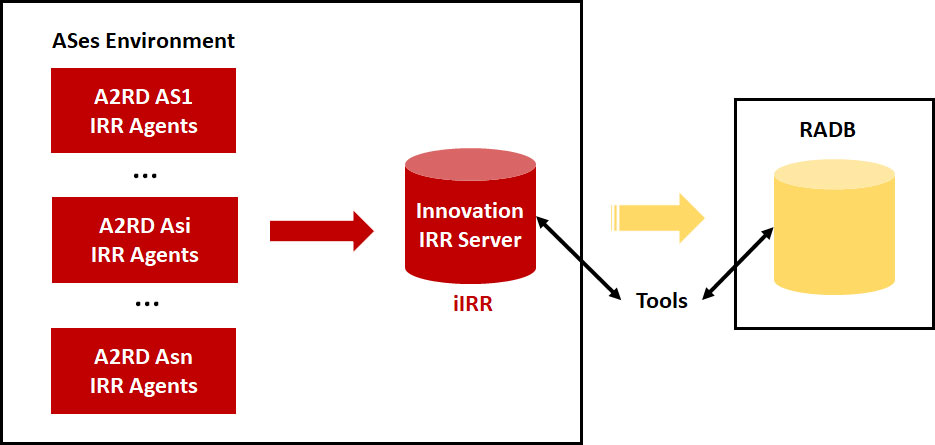}
\caption{The innovation IRR model established by A2RD}
\label{fig:innovationIRR}
\end{figure}

Sometimes the human being forgets to update information, especially those related to routing policy and that reside on important servers such as \textit{Internet Routing Registry}\footnote{http://www.irr.net/} (IRR). The IRR is a distributed database of route and route-related information \cite{Braga2010}. This fragile participation of the human being in construction and maintenance of IRR objects was the motivation for creating a model of agents that would replace human interventions on such objects. So, was implemented the \textit{Autonomous Architecture Over Restricted Domains} (A2RD) into the domain of an AS, applying as use case over the IRR \cite{Braga:2015}. A2RD replaces the human with your agents, \textit{Intelligent Elements} (IEs), establishing a new IRR model, named \textit{innovation IRR} (iIRR), shown in Figure \ref{fig:innovationIRR}.   

A special A2RD IEs, named specialized IEs, automatically create objects as defined by the \textit{Route Policy Specification Language} (RPSL) \cite{RFC2622, RFC4012}. Those objects that can not be created automatically will receive support from AS administrators through a human-computer cooperation mechanism. Nothing is changed in relation to the present and future IRR structure, characterized by the expectations recommended by the stakeholders to the \textit{Internet Engineering Task Force}\footnote{https://ietf.org/} (IETF) and \textit{Internet Research Task Force}\footnote{https://irtf.org/} (IRTF) disseminated through of yours formal documents \cite{RFC2650, RFC2725, RFC3707, RFC7682, RFC7909}. Neither does it affect the security concerns surrounding the IRR and Internet governance \cite{Kuerbis:2017}. Similarly, tools that use IRR databases can be used without any modification. A very useful, among others, is the IRR Powertools\footnote{https://github.com/6connect/irrpt} (IRRPT).

The purpose of this article is to summarize the A2RD project and it is complementary to \citeonline{Braga:2018c}. Divided into four sections, the first is this introduction. The second section is a description of the A2RD abstract model. The third section shows the A2RD implementation model and the fourth section ends the article, with the conceptual model of the A2RD development environment.

\section{The A2RD Abstract Model}

\begin{figure}[!ht]
\centering
\includegraphics[scale=0.5]{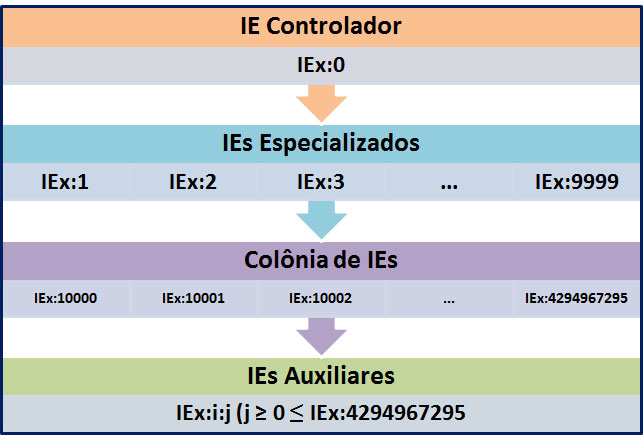}
\caption{A2RD layer model for the ASx domain, where x is the AS number.}
\label{fig:A2SD-ModeloAbstrato_EN}
\end{figure}

A2RD is a project that proposed that proposes a model of Intelligent Elements (agents) divided into layers as can be seen in Figure \ref{fig:A2SD-ModeloAbstrato_EN}.

The model serves the interest of establishing an architecture of intelligent elements on the administrative domain of ASs. It may exist in any of the \textbf{$2^{32}$} possible ASs. However, on 25/08/2018 there were only \textbf{61612} ASs, originating traffic on the Internet, according to CIDR-Report\footnote{\url{http://thyme.rand.apnic.net/current/data-summaryl}}. The number of an AS is unique, controlled by the \textit{Public Technical Identifiers}\footnote{\url{https://pti.icann.org/}} (PTI) and is named \textit{Autonomous System Number} (ASN). Thus, the largest possible value of \textbf{x} is \textbf{61612}, corresponding to \textbf{AS61612}, at the date above. A2RD implementations are independent and restricted to an AS, but with a high degree of interoperability and, of course, intensive cooperation because AS administrators depend on the behavior of all others. The PTI has reserved two contiguous ranges of AS numbers for private use: \textbf{64512-65534} and \textbf{4200000000-4294967294} \cite{RFC6996:2013}. Conveniently, these AS numbers can be used to designate Intelligent Element domains.

The first of the four layers hosts the Intelligent Element called the Controller. Its identification is unique and fixed: \textbf{x:0}, that is, the number \textbf{0} placed to the right side of the \textbf{:} symbol, following the ASN hosting the model. Sometimes, to make clear which IE is being referenced, \textbf{IE} is used before identification, for example, by stating that the \textbf{IE Controller} is \textbf{IEx:0}. Thus, if  \textbf{ASn} is the host domain of the model, then the controller element is \textbf{IEn:0}. No IE from the lower layers may exist without the prior consent of the \textbf{IE Controller}. It has the property of keeping himself organized (self-organization) and of ensuring the self-organization of any IE from the lower layer.

The second layer is represented by the so-called \textbf{Specialized IEs}. These elements are identified by suffixes that can range from \textbf{1} to \textbf{9999}. The specialized elements support the \textbf{IE Controller} in specific activities required for functionalities ranging from ensuring the interoperability of the entire system of implemented IEs to specific functionalities such as servers with end-to-end characteristics\footnote{Recognized as end-to-end arguments} that stimulate the understanding between two architectures: the layer model and the topological model \cite{Saltzer:1984}, access features to bank semantic repositories, proprietary software (similar to Southern SDN APIs), facilities required for lower-tier IEs, and many others. However, support for the \textbf{IE Controller} is the primary objective of the \textbf{Specialized IEs}. This objective is what determines the features of the second layer. It is assumed that some \textbf{Specialized IEs} may be \textbf{Autonomic Elements} or intelligent elements that execute automatic processes, such as proprietary software and procedures associated with legacy systems, among others. A \textbf{Specialized IE} can be created with functions that only concern the \textbf{IE Controller}, especially when it depends on the functionalities of IEs of the third layer.

In the third layer lies the largest agglomeration of IEs, which is why it is called the \textbf{IE Colonies}. Elements of this layer can be \textbf{autonomous},\textbf{autonomic} or \textbf{automatic}, except \textbf{legacy} and are directly responsible for the most important activities of the application, including software reuse. They act under the influence of a high degree of interoperability and cooperation between them and between IEs of other layers and other domains / subdomains. They do not directly participate in interconnections or exchange messages with other IEs outside the domain, but they do so through IEs in the upper tiers. There is intense semantic interoperability activity on the part of these IEs, which have a high capacity for self-learning due to continuous interactions with the domain environment, and produce improvement effects on the knowledge of other IEs of the \textit{colony} itself and the IEs of the layers the \textbf{IE Controller}. In other words, these IEs favor the learning of the entire cluster of IEs of the layer model, which is being described. The IEs of the colonies receive an identification with numerical suffixes, ranging from \textbf{10000} to \textbf{4294967295}.

The fourth layer is the \textbf{Auxiliary IEs}. This layer exists, in order to allow the transfer of computing demands to a new set of IEs (A2RD successiveness). It reproduces, successively, the first, second, third and fourth layers. This new IEs sequence has an additional suffix \textbf{:j:0} for a new \textbf{IE Controller} responsible for the next four layers. In the second, third and fourth new layers, the IEs identifications are postfixed with \textbf{:j:id} where, \textbf{j} is the colony IE number that originated the new fourth layer and \textbf{id} is a number with the above specifications. A typical application for the fourth layer are subdomains, such as home networks (\textbf{homenet}).

The use case for the A2RD was the addition and update of objects in IRR server. The application was considered useful mainly because the tasks of the AS administrator did not guarantee the accuracy in its completion nor the permanent need to update the objects making the IRR an unreliable system from the point of view of its contents. A2RD solved this problem \cite{Braga:2017}. 

\section{The A2RD Implementation Model}

\begin{figure}[!ht]
\centering
\includegraphics[scale=0.6]{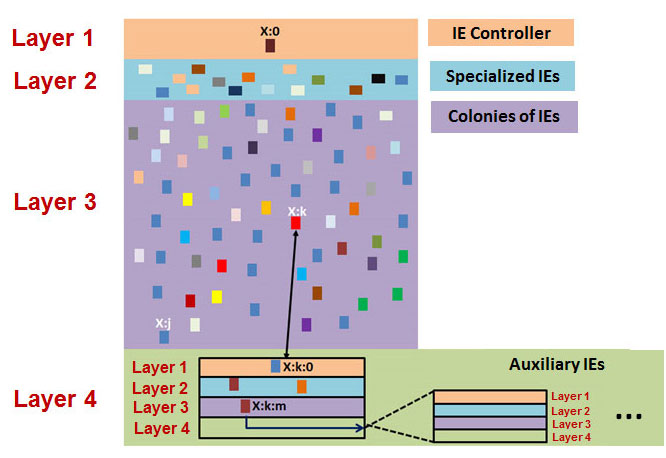}
\caption{A2RD implementation model, where x is any ASN.}
\label{fig:A2RD-ModeloImplementacao1_E}
\end{figure}

According to Figure \ref{fig:A2RD-ModeloImplementacao1_E}, the IEs are arranged and distributed between layers, similar to what was said in the previous section and are implemented in the domain of any ASN.

It is also observed, in the same figure, that the IEs functionally important in inter-domain operations reside in the upper layers. For example, a classification of relevance is the intensity of aggregation that an IE possesses, in relation to the auto-* (or self-*) properties. If an IE, however, has some self-organizing capability, it must participate directly linked to the \textbf{IE Controller}. Even if it participates in the \textbf{Auxiliary IEs} layer, an \textbf{IE Controller} can logically construct a new layer architecture. And so on.

On the other hand, the representation of the model is logical (abstraction of the physical implementation). Physically, the locating if an IE in the domain environment is essential. The best alternative is IP addressing, preferably IPv6, for reasons of availability. The IE Controller must maintain a table associating the reference reference logic with the IP designated by the IE Controller itself, from the premise that an IPv6 block should be available at the beginning of the implementation. In the implementation of prototypes related to the case study, the Python language will be used. When needed, features closer to the operating system will be used ("scripts" and other inherent facilities). 

\section{The A2RD Environment Conceptual Model}

\begin{figure}[!ht]
\centering
\includegraphics[scale=0.3]{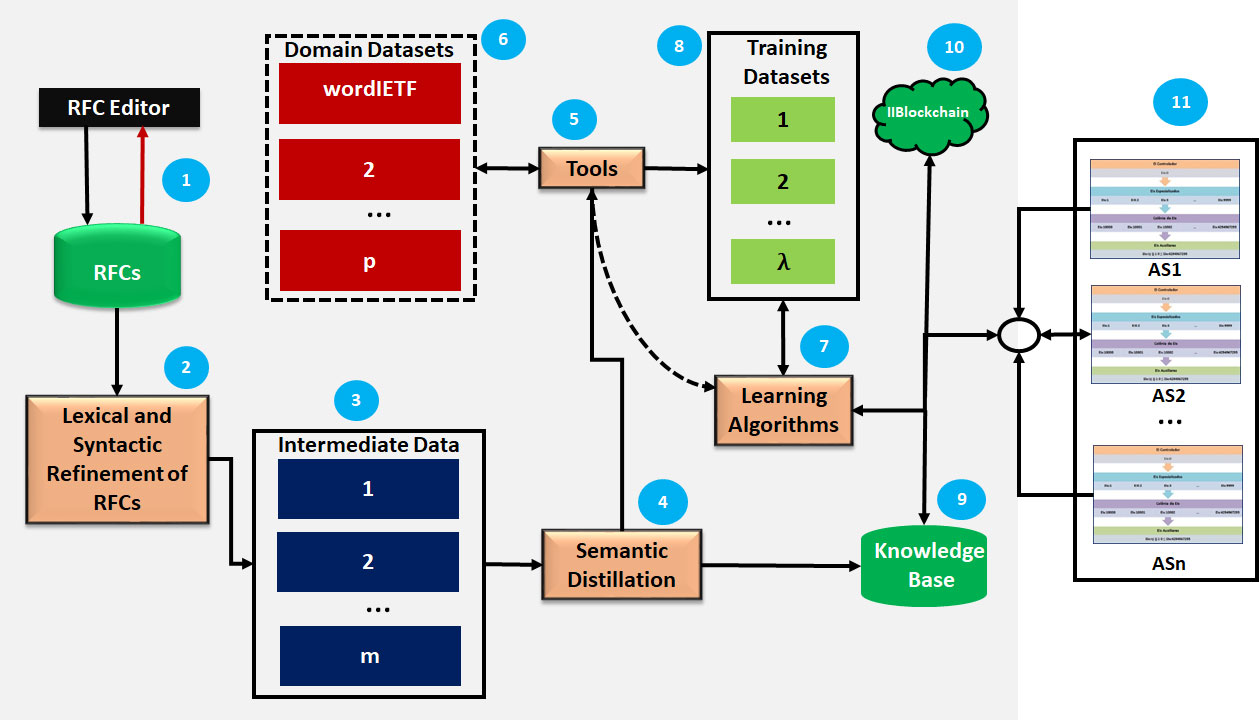}
\caption{Structure for Knowledge Acquisition and Use (SKAU)}
\label{fig:skau}
\end{figure}

The Figure \ref{fig:skau} shows the \textit{environment conceptual model}, named \textit{Structure for Knowledge Acquisition, Use and Collaboration Inter A2RD Agents} (SKAU) in which each implementation of A2RD, into an AS, is represented as an agglomeration of IEs in a four layers model (11).

 The other components of the SKAU are dynamically constructed from non-structured databases, in this experiment, from the Request for Comments (RFCs) database containing documents authored by network operators, engineers and computer scientists, documentary methods, behaviors, research, or innovations applicable to the Internet, all of them, working in groups of the IETF and IRTF, and maintained by RFC-Editor\footnote{https://www.rfc-editor.org}.

These SKAU components can be described as following:

\begin{itemize}[topsep=0pt,itemsep=1ex,partopsep=1ex,parsep=1ex]
	\item RFCs are captured / updated and stored locally (1);
	\item A set of tools responsible for acting lexically and syntactically on RFCs (2), transforming them into intermediary databases (3);
	\item Other tools (4), like \textit{Semantic Distillation}, that act on the intermediary databases producing inputs for the construction of \textit{Domain Datasets} (6) and so these into \textit{Training Data Sets} (8). Also, these tools will support for provide part of the \textbf{knowledge base} (9) \cite{isotani2015};
	\item \textit{Learning algorithms} (7) that support the construction and use of \textit{Training Datasets} to renew the knowledge base and meet the demand of agents of A2RD models in the process of developing and applied intelligent actions.
	\item A database, named \textit{IIBlockchain} (10) built by each implemented A2RD model and stored together in the Git Hub (so, in cloud), that serves as support for the process of collaboration and effective interaction, inter / intra agents of the models \cite{Braga:2018}. The \textit{IIBlockchain} cloud interacts with the \textit{learning algorithm} and \textit{knowledge base} allowing agents to exercise \textit{offline and online computation}\footnote{\textit{Offline computation is the computation done by the agent before it has to act, and online computation is the computation done by the agent between observing the environment and acting in the environment} \cite{poole2010artificial}}.
\end{itemize}  

Each AS can implement an A2RD, which is controlled by the IE named \textit{IE Controller}, and receives the identification \textit{x:0}, where \textit{x} is the \textit{AS Number} (ASN).

\section{Thanks}

From Juliao Braga: Supported by CAPES – Brazilian Federal Agency for Support and Evaluation of Graduate Education within the Brazil’s Ministry of Education and was also supported by national funds through Fundação para a Ciência e a Tecnologia (FCT) with reference UID/CEC/50021/2013.

\bibliographystyle{sbc}
%\bibliography{\bib}  % Comentar para o arXiv
\bibliography{pdbib}

\end{document}